# KNN Ensembles for Tweedie Regression: The Power of Multiscale Neighborhoods

Colleen M. Farrelly

Independent Researcher (Past, University of Miami)

**Abstract:**

Very few K-nearest-neighbor (KNN) ensembles exist, despite the efficacy of this approach in regression, classification, and outlier detection. Those that do exist focus on bagging features, rather than varying k or bagging observations; it is unknown whether varying k or bagging observations can improve prediction. Given recent studies from topological data analysis, varying k may function like multiscale topological methods, providing stability and better prediction, as well as increased ensemble diversity.

This paper explores 7 KNN ensemble algorithms combining bagged features, bagged observations, and varied k to understand how each of these contribute to model fit. Specifically, these algorithms are tested on Tweedie regression problems through simulations and 6 real datasets; results are compared to state-of-the-art machine learning models including extreme learning machines, random forest, boosted regression, and Morse-Smale regression.

Results on simulations suggest gains from varying k above and beyond bagging features or samples, as well as the robustness of KNN ensembles to the curse of dimensionality. KNN regression ensembles perform favorably against state-of-the-art algorithms and dramatically improve performance over KNN regression. Further, real dataset results suggest varying k is a good strategy in general (particularly for difficult Tweedie regression problems) and that KNN regression ensembles often outperform state-of-the-art methods.

These results for k-varying ensembles echo recent theoretical results in topological data analysis, where multidimensional filter functions and multiscale coverings provide stability and performance gains over single-dimensional filters and single-scale covering. This opens up the possibility of leveraging multiscale neighborhoods and multiple measures of local geometry in ensemble methods.

Introduction

Tweedie-distributed outcomes are ubiquitous in data analysis, spanning fields such as actuarial science (DeJong & Heller, 2008; Freese et al., 2016; Jorgensen & Paes De Souza,1994; Manning et al., 2005;), engineering (Miaou et al., 1992; Montgomery et al., 2009), psychology/medicine (Alhusen et al., 2013; Fehm et al., 2008; Ferguson et al., 2006; Herings & Erkens, 2003; Kendal et al., 2000; Mills et al., 2006; Moger et al., 2004; Wang et al., 2015), ecology (Bulmer, 1974; Candy, 2004; Hasan & Dunn, 2011; Luitel, 2016; Meng et al., 2007; Villarini et al., 2016), and criminology (Land et al., 1996;), and these outcomes can be modeled by generalized linear models. Distributions within the Tweedie framework include

Poisson, negative binomial, gamma, zero-inflated Poisson, compound Poisson-gamma, inverse Gaussian, extreme stable, and normal distributions; these distributions can also include underdispersion and overdispersion, where variance may be less or more than predicted by the model, respectively (Gardner et al., 1995; Jorgensen et al., 1994; Tweedie, 1984). As such, Tweedie models are extremely flexible and have the potential to model many outcomes of interest in academic and industrial research.

An influx of machine learning methods that can handle a variety of generalized linear modeling links have proven to be versatile new tools, easily capturing interaction terms/main effects in the model and handling high-dimensional data or data with predictors/errors violating assumptions of generalized linear modeling. Algorithms such as random forest and boosted regression employ ensemble techniques (bagging and boosting) to create stable predictors out of unstable baselearners (Breiman, 2001; Friedman, 2002). Boosted regression models can also include regression penalties, which impose sparsity and induce robust estimation similar to elastic net; a common implementation of this framework is xgboost, which has seen a lot of success on Kaggle competitions in recent years (Chen & Guestrin, 2016).

Other machine learning methods, such as neural networks, rely on a series of topological mappings to process the data and minimize prediction error between those mappings and an outcome. Extreme learning machines (ELMs), which employ a single, wide layer of hidden nodes and random mapping of features to hidden nodes, have seen success on many regression and classification problems in recent years, and these often outperform other machine learning methods (Huang et al., 2006).

One topologically-based machine learning model, Morse-Smale regression, relies upon first partitioning data by defining Morse-Smale complexes using a Morse function (typically the outcome of interest in the regression) and then fitting a piece-wise elastic net model to the individual partitions (Gerber et al., 2013). Because of its reliance on neighborhood selection through a KNN graph, Morse-Smale regression provides a nice bridge between other machine learning models and KNN methods. Though few applications exist in the literature, this approach has nice theoretical guarantees that suggest its potential for modeling continuous outcomes that fit within a generalized linear model framework, such as Tweedie regression.

K-nearest-neighbor (KNN) regression, a nonparametric regression method, has also been used to solve generalized linear modeling problems (Altman, 1992), particularly for classification problems (Fuchs et al., 2016; Hwang & Wen, 1998); KNN methods have also been successfully used on distribution-based problems such as outlier detection (Divya & Kumaran, 2016; Hautamaki et al., 2004)). One drawback of the method is its inability to deal with high-dimensional and sparse data, known as the curse of dimensionality (Aggarwal & Yu, 2001; Hastie et al., 2001; Kramer 2011). Because of this, locally-based extensions, dimensionality reduction, and random-feature-based ensembles have been developed to overcome this limitation (Bay, 1998; Dominiconi et al., 2000; Dominiconi & Yan, 2004; Johansson et al., 2010; Suguna & Thanushkodi, 2010). Despite some success with ensembles of KNN regression, very few examples exist in the literature, and all focus on bagging of features, rather than exploring the role of varying the k parameter or bagging individuals (which would alleviate the computational burden for large datasets). The importance of choosing a good k for optimal algorithm performance has been noted in many studies (Bay, 1998; Dominiconi & Yan, 2004), including those developing feature-bagged ensembles, but this has not been varied in any existing KNN ensembles. Important evidence from recent theoretical work on topological data analysis methods suggest that varying topological neighborhood size could allow a KNN ensemble to capture multiscale features and provide added stability to the algorithm (Carlsson & Zomorodian, 2009; Cerri et al., 2013; Dey et al., 2016; Dey et al., 2017). In addition, the computational limitations inherent in KNN problems has be noted but not investigated within a bagged observation framework; the benefits of designing parallelizable ensembles for MapReduce computing frameworks and of reducing training sample size through bagging observations has been noted many times (Han et al., 2013; Li et al., 2012).

This study aims to develop a series of KNN regression ensembles based on varying k, bagging features, and bagging observations to understand how each of these strategies and their combinations can improve KNN regression methods for Tweedie outcomes, as well as how these ensembles can perform relative to state-of-the-art algorithms like random forest, boosted regression (adaboost and xgboost), Morse-Smale regression, and extreme learning machines. Performance is tested and compared on 36 low- and high-dimensional Tweedie regression simulations with a variety of dispersions, varied Tweedie parameters, and linear/nonlinear predictor relationships, as well as 6 real datasets. The real datasets include problems predicting gas mileage, forest fire size, ozone level, housing price, insurance claim payouts, and material strength.

It is hypothesized that KNN regression ensembles will perform comparably well and that KNN regression ensembles with randomized k, feature subsets, and observation subsets will perform well with the added benefit of lowered computational cost for each baselearner in the ensemble (and, thus, the best computational performance within a MapReduce framework or on a laptop/desktop without parallel computing).

**Methods**

**I) Tweedie Regression**

Tweedie distributions belong to the exponential family, in which links can be defined to generalize linear regression to non-normal outcomes, called generalized linear models (Tweedie, 1984); their distributional parameters allow for the flexible modeling of count and ratio data with small and large variances within a sample (Bonat et al., 2016). Many common distributions of the exponential family converge to Tweedie distributions and can be formulated through Tweedie distributions, and Tweedie distributions themselves enjoy a variety of useful properties, including reproductive properties that allow distributions to be added together to form new distributions that are themselves Tweedie (Jorgenson et al., 1994; Tweedie, 1984). This allows many regression problems to be cast as Tweedie regression (Aalen, 1992). Formally, the mean and variance are given by:

$$E(Y) = \mu$$

$$Var(Y) = \varphi\mu^\xi$$

where $\varphi$ is the dispersion parameter, and $\xi$ is the Tweedie parameter (or shape parameter). A Tweedie parameter of 0 corresponds to the normal distribution; 1 to the Poisson distribution; and 2 to the gamma distribution. Compound Poisson-gamma distributions have a Tweedie parameter between 1 and 2, and these have been quite useful in modeling biological data. Tweedie parameters greater than 2 generally yield stable distributions.

Given the wide variety of data modeling problems that involve an outcome which can be modeled through the Tweedie distribution, it is likely that algorithms performing well on a variety of Tweedie regression tasks will perform well on many generalized linear regression tasks (and, thus, would represent a good general solution to regression tasks).

**II) A Few State-of-the-Art Machine Learning Regression Methods**

Several techniques, including other ensemble methods, were used as baseline comparison on simulations and real data analyses. Random forest is a bagged ensemble of tree predictors, with each tree grown on a bootstrap sample of features and observations, where results are pooled from all bootstrap models (Breiman, 2001). It is highly effective at creating diversity within the ensemble to outperform single tree methods, and it performs well across a variety of classification and regression problems (Fernandez-Delgado et al., 2014).

Boosted regression models are created by iteratively fitting and averaging models, typically through gradient descent methods; subsequent models are fit to model error terms, which promote diversity through fitting specific model pieces not captured in previous iterations (Friedman, 2002). Baselearners used in the ensemble can be linear pieces, trees, or other models, offering flexibility and increasing the potential for diversity within the ensemble (Hastie et al., 2001). One recent extension that has performed well on Kaggle data science competitions is xgboost (Chen & Guestrin, 2016), which introduces penalties to the boosted regression equations much like those used in elastic net models (ability to set both $\ell_1$ and $\ell_2$ penalties to induce model sparsity and estimate robustness, respectively).

Morse-Smale regression is a piecewise elastic net model based on decomposing the model space through the use of topology defined by KNN-based local neighborhoods (Gerber et al., 2013). Morse functions, a special type of continuous function analogous to a height function used in topography, are used to explore critical points of a data manifold and their basins of attraction. These basins of attraction serve as a way to partition the data manifold into pieces with common local minima and maxima flow (see Figure 1, Chen et al., 2017); regression models are then fit to these pieces. Though not technically an ensemble method, its piecewise nature suggests the possibility for diversity of models across partitions similar to that found in common ensemble models. Little is known about its performance on real-world datasets compared to other methods, but preliminary results from the development papers suggest it could perform well on Tweedie regression problems. Its reliance on KNN-defined neighborhood kernels provides a nice link to KNN regression.

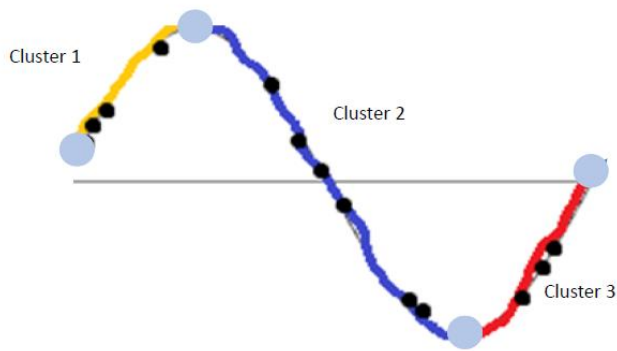

*Figure 1: Example of Morse-Smale cluster on part of a sine wave based on height filter function with basins of attraction*

Lastly, extreme learning machines are feedforward neural networks with one hidden layer and random mappings between the input space and the hidden layer (Huang et al., 2006), which allows for the use of least squares methods to optimize weights between the hidden layer and output (rather than costly algorithms like gradient descent, momentum, or adaptive fit techniques). It has performed well on classification and regression problems (Fernandez-Delgado et al., 2014), and given its universal approximation property (Huang et al., 2006), extreme learning machines have the potential for performing well across a variety of Tweedie regression problems, though actual applications thus far have been limited.

**III) KNN Ensemble Formulation**

As detailed in the aforementioned section, model diversity is a key component of creating ensembles which outperform their baselearners. In previously designed KNN ensembles (Domeniconi & Yan, 2004), this is mainly achieved through choosing random subsets of features or optimized subsets of features; after features are chosen, k is typically optimized for each subset before the models are combined into the ensemble (Johansson et al., 2010). Because k is such an important parameter and different values of k will give different regression estimates, it is likely that varying k or randomly selecting k within ensembles can produce substantial model diversity, even with the same samples or feature sets. In this way, varying k varies the topological neighborhood defined around data points, providing different lenses through which to view predictor-outcome relationships and smoothing of a regression function (see Figure 2). Ensembles with this approach essentially have randomly chosen topological neighborhoods. In addition, though KNN models are typically robust to subsampling of a population (stable with respect to sampling), bootstrapping observations within a KNN ensemble has not been investigated previously; this represents another possibility for designing a KNN ensemble, providing diversity akin to random forest when combined with feature bootstrapping and capturing multiscale sample features when combined with varying k.

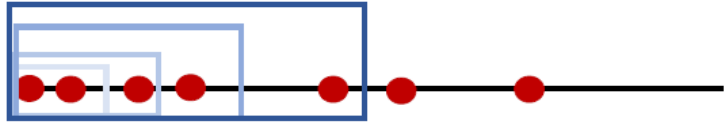

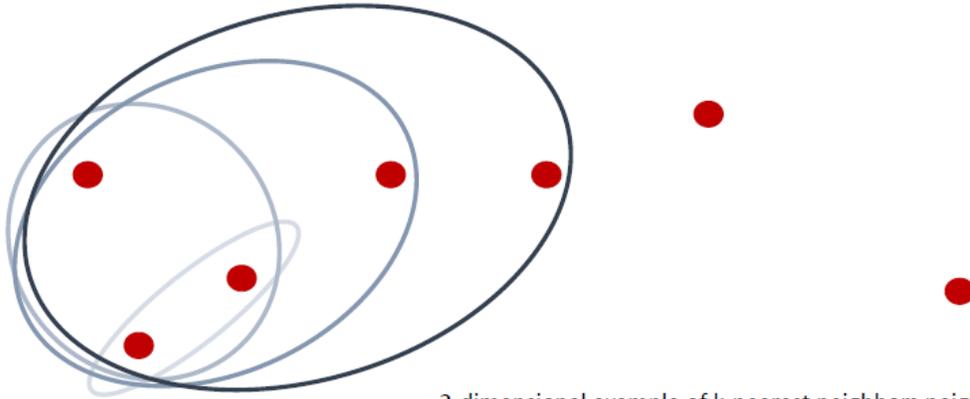

*Figure 2: Multiscale neighborhood example with varying k (1- and 2-dimensional examples)*

Given these potential variations, 7 combinations are possible, yielding 7 distinct KNN regression ensembles that can be compared on simulation data and on real-world problems. All 7 combinations utilize a bagging approach with averaged estimates and Euclidean distance metric for simplicity (though varying distance functions offers another potential for creating diverse ensembles), and a single KNN regression model with tuned k is used for comparison in benchmark tasks. All ensembles were optimized with respect to bagging fractions and k across simulations, with the best parameters across problems selected for use in the ensembles, such that the best general performance of each ensemble could be compared. All ensembles grew 10 baselearner models, as this was the point at which prediction gains leveled off.

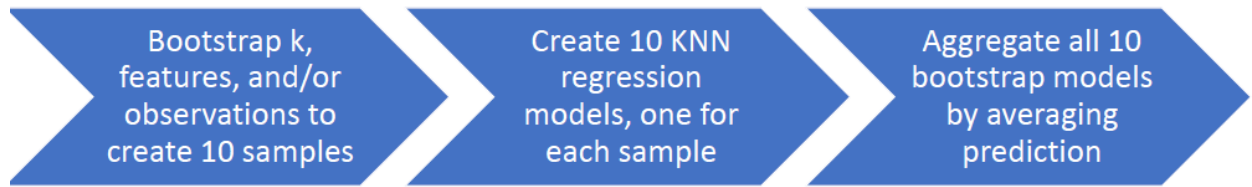

*Figure 3: General KNN regression ensemble algorithm steps*

Figure 3 shows the ensemble construction steps for the 7 derived ensembles:

1) Vary samples and features selected (30% of observations, 90% of features, k=10)

2) Vary samples, features, and k (30% of observations, 90% of features, k randomly chosen from 5 to 15)

3) Vary features (30% of features, k=10)

4) Vary k (chosen randomly between 5 and 15)

5) Vary samples and k (30% of observations, k randomly chosen from 5 to 15)

6) Vary samples (30% of observations, k=10)

7) Vary features and k (30% of features, k randomly chosen from 5 to 15)

One advantage of this formulation is the ability to run these algorithms within a MapReduce framework to ameliorate the computational burden for large datasets. Thus, an ensemble with 10 baselearner models could be run on 10 cores and have the results pooled to create a final prediction. Bootstrapping of observations (varying samples) reduces computational burden, as well, for long datasets (such as large insurance datasets); bootstrapping features reduces computational burden for wide datasets (such as genomics or healthcare datasets with lots of predictors). Thus, KNN regression ensembles with random k, random feature subsets, and random observation sampling should have the lowest computational burden and potentially greater diversity than the other ensembles.

## IV) Simulations and Real Data Applications

Simulations were set up in R by varying combinations of several factors important to assessing performance of KNN ensembles. For the first round of simulations, 4 true predictors and 6 noise variables were simulated with different predictor relationships to the outcome (linear only, nonlinear only, and mixed), different outcome overdispersion levels (1, 2, and 4—no overdispersion to fairly high overdispersion), and different outcome Tweedie parameters (1, 1.5, and 2 corresponding to Poisson, compound Poisson-gamma, and gamma distributions). A constant level of noise was added to each dataset to represent measurement error and other sources of error common to real datasets; the parameters added were Binom(5,0.1), so as to include potential outliers. Each trial consisted of 10 replicates, sample sizes of 500, 1000, 2500, and 5000 for comparison of convergence properties, and a 70/30 train/test split to assess performance. All models were scored for performance using a derived model $R^2$ value:

$$R^2 = 1 - \frac{Square\ error\ of\ model}{Square\ error\ of\ mean\ model}$$

To assess the extent to which KNN ensembles suffer from the curse of dimensionality, 9 trials were simulated with 1500 noise variables and 1000 observations. Each trial consisted of a combination of predictor relationships (linear, nonlinear, mixed), overdispersion levels (1, 2, 4), and Tweedie parameter (1, 1.5, 2); these were replicated 10 times with a 70/30 train/test split and assessed with the $R^2$ value.

Algorithms were also compared on 6 real datasets from on-line repositories (70/30 train/test splits and mean square error to assess performance relative to the mean model). The first dataset, UCI Repository's Auto MPG dataset, consisted of predicting miles per gallon of each car using 7 predictors (cylinders in the car, displacement, horsepower, weight, acceleration, model year, origin, and car name) and 398 instances.

The second dataset, UCI Repository's Forest Fires dataset, consisted of predicting hectares burned in a wild fire using 12 predictors (2 spatial coordinates of location, month, day, FFMC index, DMC index, DC index, ISI index, temperature, relative humidity, wind, and rain) on 517 observations; this is a notoriously hard problem on which most algorithms do no better than the mean model.

The third dataset, the Ozone dataset from the mlbench R package, aimed to predict ozone reading using 12 predictors (month, day, day of week, millibar pressure at Vandenberg Air Force Base, wind speed at LAX, humidity at LAX, temperature at Sandburg, temperature at El Monte, inversion base height at LAX, pressure gradient from LAX to Daggett, inversion base temperature at LAX, and visibility at LAX) and 366 observations.

The fourth datasets, the Boston Housing dataset from the mlbench R package, aimed to predict median house value using 13 predictors (crime per capita, proportion of land zoned for lots >25000 square feet, proportion of non-retail business acres, Charles River tract, nitric oxide concentration, number of rooms per house, proportion of units built prior to 1940, distance from 5 major employment areas, accessibility to radial highways, property tax rate, pupil-teacher ratio in town, proportion of African-Americans by town, and percentage of lower-status population of town) and 506 observations.

The fifth dataset, the Swedish 3$^{rd}$ party motor insurance dataset from http://www.statsci.org/data/glm.html, consisted of predicting the value of 1977 payouts from insurance policies using 6 predictors (kilometers traveled a year, geographical zone, bonus, car model make, number of years insured, and total number of claims) and 2182 observations.

The final dataset, Dialectric Breakdown Strength from http://www.statsci.org/data/glm.html, predicted breakdown strength from accelerated tests with 2 predictors (duration of testing and temperature) and 128 observations.

## Results

### I) Simulations (n>p)

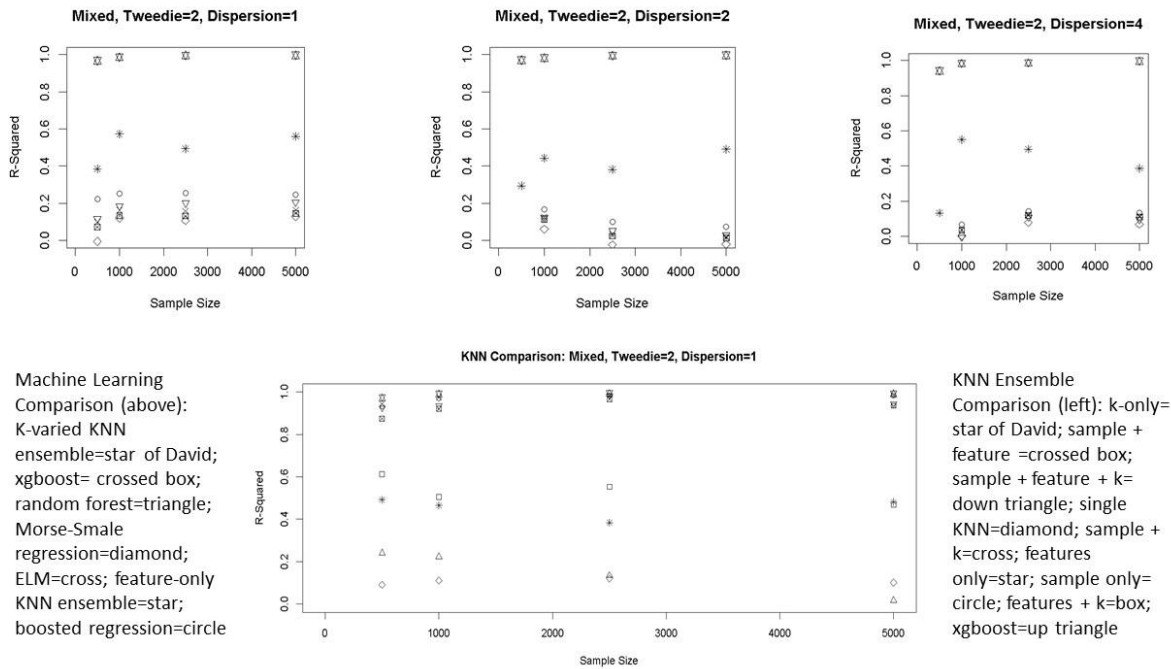

*Figure 4: N>p comparison with representative results*

A few general trends emerged and held across trials (Figure 4). Tweedie parameter seemed to have a negligible effect for all models tested, but the performance of machine learning methods (including the single KNN regression model) degraded with increasing overdispersion, suggesting that many existing methods do not generalize well to data with overdispersion (including zero-inflated data and data including outlier populations with extremely high outcome values). However, the KNN regression ensembles were robust to overdispersion.

With respect to the KNN regression ensembles, many performed extremely well across trials, exceeding an $R^2$ of 0.9; this suggests that KNN regression ensembles capture most variance with respect to a Tweedie-distributed outcome. However, two models performed significantly worse than the other KNN regression ensembles across all 27 trials, with $R^2$ values generally between 0.4 and 0.6. These were the feature-only ensemble, which is the basis for most existing KNN ensembles, and the feature subset with random choice of k. This suggests that the existing paradigm for creating KNN ensembles is not ideal and can be substantially improved.

Of the best KNN regression ensembles, the one bootstrapping observations and features with varying k gave the best computing performance, suggesting that this may be an effective strategy for high performance computing with Tweedie-distributed outcomes.

## II) Simulations (p>n)

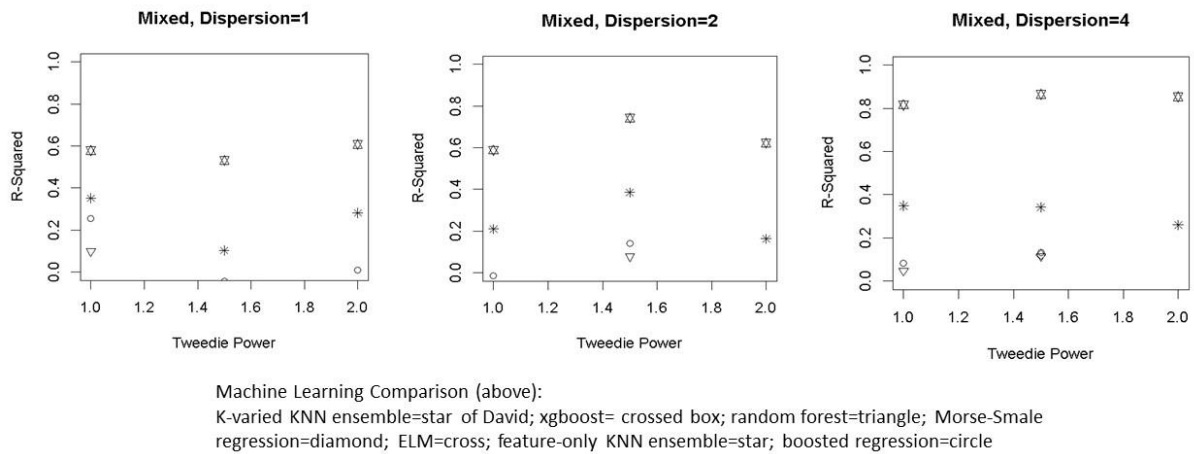

Machine Learning Comparison (above):
K-varied KNN ensemble=star of David; xgboost= crossed box; random forest=triangle; Morse-Smale regression=diamond; ELM=cross; feature-only KNN ensemble=star; boosted regression=circle

*Figure 5: P>n trials with representative results*

Adding 1500 noise variables to the models with 1000 observations—representing challenging Tweedie regression problems with sparsity and high dimensionality—proved a challenge to most machine learning algorithms, including the single KNN regression (Figure 5). The KNN regression ensembles also suffered under sparse conditions and high dimensionality; however, they did maintain significantly better performance than the other machine learning models. This suggests that while KNN regression ensembles suffer from the curse of dimensionality to some extent, they are also fairly robust to this phenomenon compared to other machine learning models and non-ensemble KNN regression models. The feature-only ensemble and feature subset with random k ensemble both showed lower $R^2$ values (~0.2 to 0.4) than the remaining KNN regression ensembles (~0.6 to 0.8), again demonstrating that the existing framework with KNN ensembles can be improved.

Another interesting trend emerges with respect to the high-performing ensembles. Uniformly, these models see dramatic increases in $R^2$ values with increasing overdispersion. Because many of these methods are based on varying k and/or selecting different observation subsets, this trend may be a result of neighborhood properties being captured more effectively under varying observation sets or varying neighborhood sizes. Given the relatively poor performance of feature subsets with random k, it is likely that the bootstrapping of observations underlies this phenomenon. An alternative may be that feature subsets somehow do not provide enough salient neighborhood characteristics for varying neighborhoods to adequately capture these characteristics in the ensemble.

**III) Real Datasets**

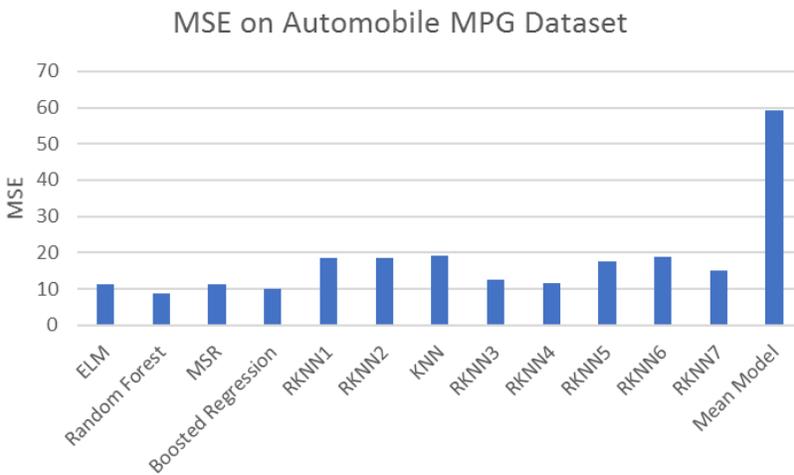

*Figure 6: Auto MPG Dataset Results*

On the Automobile MPG dataset, most algorithms performed significantly better than the mean model, with machine learning models generally outperforming KNN regression ensembles (Figure 6); the features-only and random-k-only ensembles were the exceptions, performing on par with the main machine learning models. The remaining ensembles matched or exceeded the single KNN regression model, suggesting that ensemble methods do improve KNN regression.

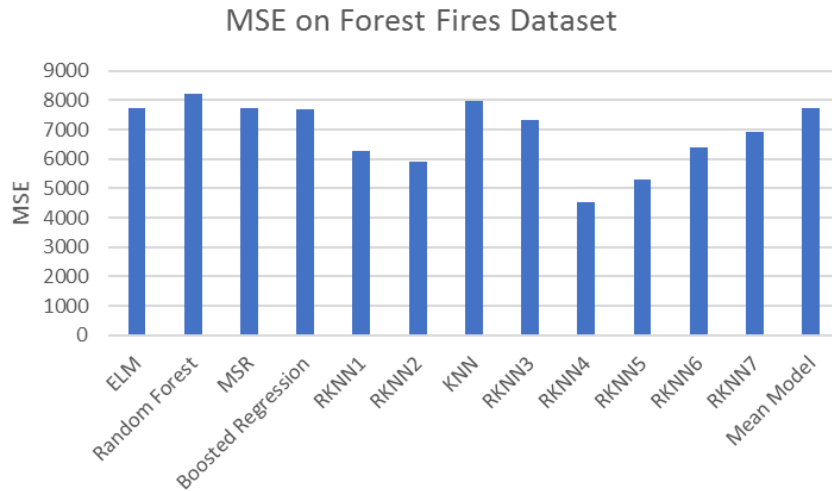

*Figure 7: Forest Fire Dataset Results*

On the Forest Fires dataset, most algorithms struggled to beat the mean model (Figure 7). Notable exceptions included KNN regression ensembles utilizing observation bootstrapping and the random-k-only model, which performed significantly better than any other model (capturing roughly half of the outcome variance). This suggests that varying k is an effective strategy, and this particular dataset may have multiscale characteristics that are missed by models with an optimal neighborhood size rather than multiple neighborhood sizes within an ensemble, echoing recent theoretical results in topological data analysis (Dey et al., 2017).

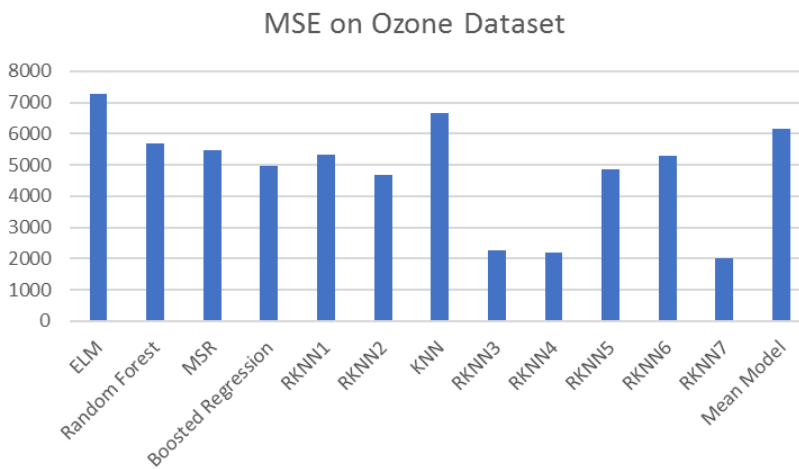

*Figure 8: Ozone Dataset Results*

On the Ozone dataset, most algorithms again struggled to beat the mean model (Figure 8). Three KNN regression ensemble methods manage to not only beat the mean model but capture over 67% of the outcome variance. These include the random-k-only model and two feature-subset with random k models, suggesting that varied neighborhood size is again important in successfully capturing salient predictor relationships and local characteristics. This also suggests feature-based bootstrapping can be very effective in certain Tweedie regression problems.

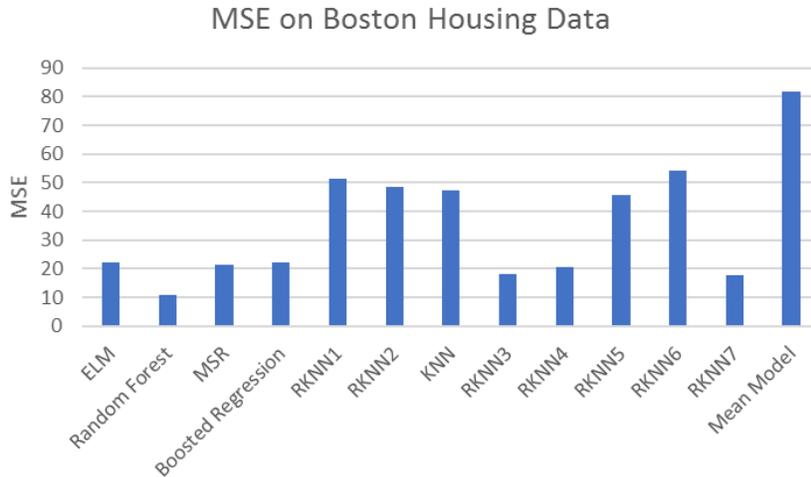

Figure 9: Boston Housing Dataset Results

Most machine learning models, particularly random forest, and the same three KNN regression ensembles that performed well on the Ozone data performed well on the Boston Housing dataset (Figure 9). The random forest model posted near perfect prediction, suggesting that many complex interaction terms exist. This is supported by the relatively good performance of feature-based bootstrapping KNN regression ensembles; the random-k-only performance suggests again that features vary across neighborhood size.

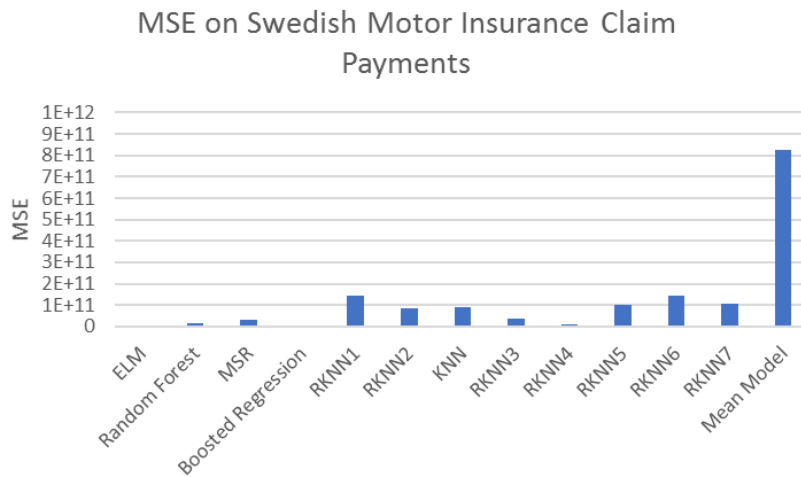

Figure 10: Swedish Insurance Claim Payout Dataset Results

Every single model performed extremely well on the Swedish Insurance Claims dataset (Figure 10), with the random-k-only KNN regression ensemble rivaling the best machine learning models. This suggests that machine learning methods in general have great potential in solving actuarial problems. However, most KNN-based models—including the features-only ensemble—performed relatively poorly compared to the other machine learning models and random-k-only KNN regression ensemble.

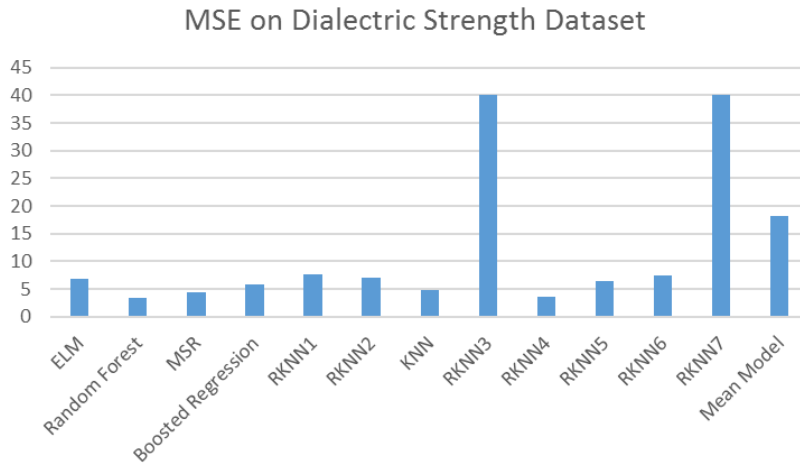

*Figure 11: Dialectric Strength Dataset Results*

On the Dialectric Strength dataset, random forest and random-k-only KNN regression ensemble both performed extremely well (Figure 11), with most other machine learning models showing good predictive capability. However, the feature-subset and feature-subset with random k models struggled to attain any predictive worth; given that this dataset has only two predictors, this likely reflects the lack of diversity and inability to train on only one predictor.

Taken together, these results suggest the efficacy of varying neighborhood size within KNN regression ensembles, as well as the potential for observation-based bootstrapping to improve KNN regression ensemble predictive capabilities beyond the extant KNN ensembles.

**Discussion**

These results provide several take-aways and tantalizing possibilities for improving machine learning algorithm performance on difficult regression problems. First, the ability of ensembles only varying k—particularly on the hardest problems—suggest that neighborhood geometry and scaled local relationships are an important aspect that is not well-captured in current machine learning algorithms, including neighborhood-based methods like KNN and Morse-Smale regression. These results echo recent work in the field of topological data analysis, where persistent homology details how features are born and die with changes of neighborhood (Carlsson & Zomorodian, 2009). This change of neighborhood captures features and aspects of features that would be missed within a single neighborhood; it captures features of different scales, as well. One result theoretical result concludes that adding multiple scales, even to parameters within persistent homology algorithms, induces stability and more powerful insight (Dey et al., 2016); thus, multiscale mappers and multiscale filter functions in persistent homology outperform algorithms utilizing single scale parameters (i.e. 2-dimensional filter functions or iteratively performing the mapper algorithm with varied covering parameters vs. 1-dimensional filter functions or a single covering parameter in the mapper algorithm). Ensembles utilizing different k parameters seem to fit into this general framework, creating better regression models than are possible with single-k regression models.

This realization offers several possible extensions. If these different neighborhoods are picking up different features and relationships in the data, building KNN regression ensembles with different distance metrics (ex. Mahalanobis, Hamming, Manhattan, even Riemannian geodesics defined on the data manifold or KNN graph) may improve prediction, as well (Weinberger et al., 2006). Combining a variety of metrics with a variety of neighborhoods may boost current prediction capabilities and offer even better methods for solving Tweedie regression problems. In addition, this principle can be applied to ensembles of Morse-Smale regression models with varying filter functions and neighborhood settings (a multiresolution Morse-Smale regression ensemble); this may improve this algorithm's performance over a single model, as well. A follow-up study aims to explore this possibility in greater depth.

One drawback of KNN regression ensembles is their blackbox nature, where little insight into the nature of the relationships may be gained from the model. By using subsets of predictors on larger ensembles, it would be possible to derive importance scores of individual features or subsets of features by examining model accuracy gains from inclusion of that particular variable over models that lack a given variable. Thus, a set-theoretic importance score could be derived without losing accuracy gains within an ensemble employing feature selection. With MapReduce frameworks, the computational burden of these calculations and the inclusion of large subsets of features (say, 90% of features) could be mitigated.

However, the findings of this paper are encouraging and offer a solution for improving not only KNN ensembles but also other algorithms that include neighborhood size as a parameter. By following the theoretical results in many areas of topological data analysis, a new direction for ensemble learning methods is opened. Neighborhoods matter, and varying them offers a potential to solve the more difficult problems in the field of machine learning and generalized linear modeling.